\documentclass[conference]{./support/ieeeconf}
\usepackage{times}
\usepackage{amsmath}
\usepackage{balance}
\usepackage[pdftex]{graphicx}
\usepackage{subfigure}
\usepackage{amsmath,amssymb,amsopn,amstext,amsfonts}
\usepackage{cancel}
\usepackage[space]{cite}
\usepackage{pdfsync}
\usepackage{balance}
\usepackage{color}
\usepackage{algorithm}
\usepackage{algorithmic}
\usepackage{url}
\usepackage{booktabs}
\usepackage{threeparttable}
\usepackage[linkcolor=black,citecolor=black,urlcolor=black,colorlinks=true]{hyperref}
\usepackage{multirow}
\usepackage{graphicx}
\IEEEoverridecommandlockouts

\graphicspath{{./figures/}}
\DeclareGraphicsExtensions{.pdf,.png,.jpg}
\title{\LARGE \bf Online Temporal Calibration for Monocular Visual-Inertial Systems}
\author{Tong Qin and Shaojie Shen
\thanks{All authors are with the Department of Electronic and Computer Engineering,
        Hong Kong University of Science and Technology, Hong Kong, China.
        {\tt\small tqinab@connect.ust.hk, eeshaojie@ust.hk}.
      This work was supported by the Hong Kong Research Grants Council Early Career Scheme under project no. 26201616.}
}

\begin{document}

\maketitle
\thispagestyle{empty}
\pagestyle{empty}

\begin{abstract}
Accurate state estimation is a fundamental module for various intelligent applications, such as robot navigation, autonomous driving, virtual and augmented reality. 
Visual and inertial fusion is a popular technology for 6-DOF state estimation in recent years.   
Time instants at which different sensors' measurements are recorded are of crucial importance to the system's robustness and accuracy. 
In practice, timestamps of each sensor typically suffer from triggering and transmission delays, leading to temporal misalignment (time offsets) among different sensors. 
Such temporal offset dramatically influences the performance of sensor fusion. 
To this end, we propose an online approach for calibrating temporal offset between visual and inertial measurements.
Our approach achieves temporal offset calibration by jointly optimizing time offset, camera and IMU states, as well as feature locations in a SLAM system.
Furthermore, the approach is a general model, which can be easily employed in several feature-based optimization frameworks. 
Simulation and experimental results demonstrate the high accuracy of our calibration approach even compared with other state-of-art offline tools.
The VIO comparison against other methods proves that the online temporal calibration significantly benefits visual-inertial systems. 
The source code of temporal calibration is integrated into our public project, VINS-Mono\footnote{https://github.com/HKUST-Aerial-Robotics/VINS-Mono}.   

\end{abstract}

\section{Introduction}
State estimation has been a fundamental research topic in robotics and computer vision communities over the last decades.
Various applications, such as robot navigation, autonomous driving, virtual reality (VR) and augmented reality (AR), highly rely on accurate state estimation. 
We are particularly interested in state estimation solutions that involve only one camera, due to its small size, low power consumption, and simple mechanical configuration.
There have been excellent results in monocular visual-only techniques \cite{davison2007monoslam, klein2007parallel, ForPizSca1405, engel2014lsd, mur2015orb, kaess2012isam2, engel2017direct}, which computed accurate camera motion and up-to-scale environmental structure. 
To solve the well-known scale ambiguity, multi-sensor fusion approaches attract more and more attention.
Many researches \cite{corke2007introduction, MouRou0704,LiMou1305,weiss2012real,lynen2013robust,SheMicKum1505,mur2017visual,forster2017manifold, LeuFurRab1306,bloesch2015robust} assisted camera with IMU (Inertial Measurement Unit), which achieved impressive performance in 6-DOF SLAM (simultaneous localization and mapping).
On the one hand, inertial measurements render pitch and roll angle, as well as scale, observable.
On the other hand, inertial measurements improve motion tracking performance by bridging the gap when visual tracking fails. 

To fuse data from different sensors, time instants at which measurements are recorded must be precisely known. 
In practice, the timestamps of each sensor typically suffer from triggering and transmission delays, leading to a temporal misalignment (time offset) between different sensor streams.
Consequently, the time synchronization of sensors may cause a crucial issue to a multi-sensor system.
For the visual-inertial system, the time offset between the camera and IMU dramatically affects robustness and accuracy.
Most visual-inertial methods \cite{SheMicKum1505,LeuFurRab1306, bloesch2015robust, mur2017visual} assumed measurements' timestamps are precise under a single clock. 
Therefore, these methods work well with a few strictly hardware-synchronized sensors. 
For most low-cost and self-assembled sensor sets, hardware synchronization is not available. 
Due to triggering and transmission delays, there always exists a temporal misalignment (time offset) between camera and IMU.
The time offset usually ranges from several milliseconds to hundreds of  milliseconds. 
Dozens of milliseconds will lead to IMU sequences totally misaligning with image stream, thus dramatically influencing the performance of a visual-inertial system.
\begin{figure}
    \centering
    \includegraphics[width=0.48\textwidth]{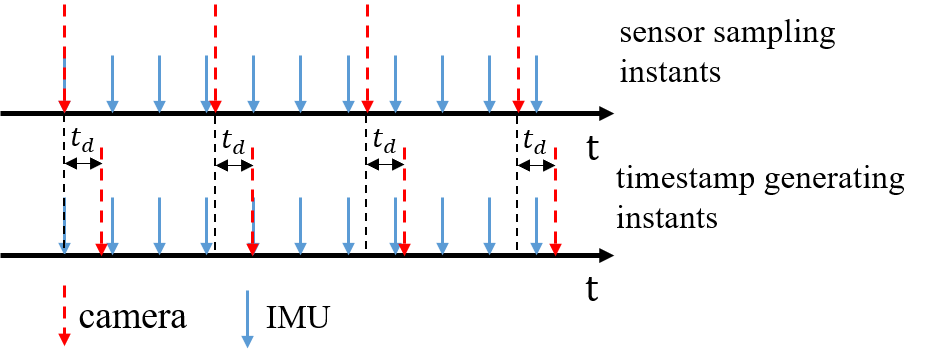}
    \caption{An illustration of temporal misalignment (time offset) between camera and IMU streams. The upper plot represents sampling instants. The lower plot shows timestamping instants.  The generated timestamp is not equal to the actual sampling time due to triggering delay, transmission delay, and unsynchronized clocks, leading to a temporal misalignment between camera and IMU. 
        The time offset $t_d$ is the amount of time by which we should shift the camera timestamps so that the camera and IMU data streams became temporally consistent. 
        \label{fig:timeoffset}}
\end{figure}

To this end, we propose a method to online calibrate temporal offset for a visual-inertial system.
We assume time offset is a constant but unknown variable.
We calibrate it by estimating it online along with camera and IMU states, as well as feature locations in a SLAM system. 
Our calibration approach is a general factor, which can be easily employed in other feature-based visual-inertial optimization frameworks. 
Although we use the monocular sensor suite to showcase our method, the proposed approach can be easily applied to multi-camera visual-inertial systems.
We highlight our contribution as follows:
\begin{itemize}
    \item We propose an online approach to calibrate temporal offset between camera and IMU in the visual-inertial system.
    \item We showcase the significance of online temporal calibration through both simulation and real-world experiments.
    \item Open-source code integrated into the public project.  
\end{itemize}

The rest of the paper is structured as follows. 
In Sect.~\ref{sec:literature}, we discuss the relevant literature. 
The algorithm is introduced in detail in Sect.~\ref{sec:algorithm}.
Implementation details and experimental evaluations are presented in Sect.~\ref{sec:experiments}. 
Finally, the paper is concluded in Sect.~\ref{sec:conclusion}.

\section{Related Work}
\label{sec:literature}
Over the past few decades, there have been tremendous researches in visual-inertial odometry techniques, which aimed to compute camera motion and environment structure with high accuracy.
The popular techniques are either filter-based framework \cite{MouRou0704,LiMou1305, weiss2012real,lynen2013robust,bloesch2015robust}, or batch optimization \cite{SheMicKum1505,mur2017visual,ForCarDel1507,forster2017manifold, LeuFurRab1306}. 
Most of visual-inertial algorithms process image by extracting robust sparse features instead of operating on the dense image.
Among these works, \cite{MouRou0704,LiMou1305, ForCarDel1507} used structure-less 
vision factor, which eliminated features by projecting visual residual onto null space.
They focus more on estimating camera or IMU motion instead of feature positions.
\cite{SheMicKum1505,mur2017visual,LeuFurRab1306} selectively kept keyframes and features in a bundle, which optimized camera motion and feature together. 
All of these methods assumed IMU and camera are precisely synchronized without temporal misalignment.

%temporal calibration
The temporal misalignment between IMU and camera is a typical issue in low-cost and self-assembled devices. 
The measurement's timestamp is misaligned with actual sampling time instant due to unsynchronized clocks, triggering delay and transmission delay. 
This time offset is unknown and needs to be calibrated. 
Several pieces of research have focused on calibrating it. 
Mair \cite{Mair2011Spatio} proposed an initialization approach for temporal and spatial calibration, which used either cross-correlation or phase congruency.
This approach formulated calibration procedure in a novel and special view.
It separated calibrated variables from other unknown variables (poses, feature positions).
Therefore, it can provide a good prior without influence from other variables.
Further on, methods modeled time offset in a more precise formulation. 
Kelly \cite{Kelly2010A} aligned rotation curves of camera and IMU to calibrate time offset. 
It leveraged a variant of ICP (iterative closest point) method to gradually match two rotation curves.  
Kalibr, which came from Furgale \cite{furgale2013unified}, estimated time offset, camera motion, as well as extrinsic parameters between camera and IMU in the continuous batch optimization procedure. 
Kalibr achieved impressive performance and became a popular toolbox. 
However, these two methods operated offline with a fixed planar pattern (such as a chessboard).
The calibration pattern provided them with robust feature tracking and association, as well as accurate 3D position.
Moreover, Li proposed a motion estimation method with online temporal calibration for the camera-IMU system in \cite{li20133}. 
The time offset was calibrated in a multi-state constrained EKF framework. 
His method had a significant advantage in computation complexity, which can be used on portable mobile devices.
Compared with his method, our optimization-based algorithm outperforms in term of accuracy, since we can iteratively optimize a lot of variables in a big bundle instead of fixing linearization error early.

\section{Algorithm}
\label{sec:algorithm}
In this section, we model temporal offset in a vision factor, and online calculate it along with features, IMU and camera states in an optimization-based VIO framework.

We briefly denote frame and notation as follows. 
$(\cdot)^w$ denotes global frame.
$(\cdot)^c$ denotes local camera frame.
($\mathbf{R}^w_c$, $\mathbf{p}^w_c$) is camera pose in the global frame, which can transform 3D feature from camera frame to global frame.

\subsection{Temporal Offset}
For low-cost and self-assembled visual-inertial sensor sets, camera and IMU are put together without strict time synchronization. 
The generated timestamp is not equal to the time instant at which the measurement is sampled due to triggering delay, transmission delay and unsynchronized clocks. 
Hence, there usually exists temporal offset between different measurements.
%This temporal misalignment between sensor streams dramatically influence the performance of fusion algorithm.  
In general cases, the time offset between sensors is a constant but unknown value. 
In some worse cases, sensors are collected with different clocks and the time offset drifts along with the time.  
This kind of sensors is unqualified for sensor fusion.

In this paper, we consider the general case, where time offset $t_d$ is a constant but unknown value.  
One picture illustrating time offset is depicted in Fig.~\ref{fig:timeoffset}.
In the picture, the upper plot represents sampling instants. The lower plot shows timestamping instants.  
The generated timestamp is not equal to the actual sampling time due to triggering delay, transmission delay and unsynchronized clocks, leading to a temporal misalignment between camera and IMU. 
Specifically, we define $t_d$ as, 
\begin{equation}
t_{IMU} = t_{cam} + t_d.
\end{equation}
The time offset $t_d$ is the amount of time by which we should shift the camera timestamps, so that the camera and IMU data streams became temporally consistent. 
$t_d$ may be a positive or negative value. 
If the camera sequence has a longer latency than the IMU sequence, $t_d$ is a negative value.
Otherwise, $t_d$ is a positive value. 

\subsection{Feature Velocity on Image Plane}
To make camera and IMU data streams temporally consistent, the camera sequence should be shifted forward or backward according to $t_d$.
Instead of shifting whole camera or IMU sequence, we specifically shift features' observations in the timeline. 
To this end, we introduce feature velocity for modeling and compensating the temporal misalignment.

In a very short time period (several milliseconds), the camera's movement can be treated as constant speed motion.
Hence, a feature moves at an approximately constant velocity on the image plane in short time period. 
Based on this assumption, we compute the feature's velocity on the image plane.
  
\begin{figure}
    \centering
    \includegraphics[width=0.25\textwidth]{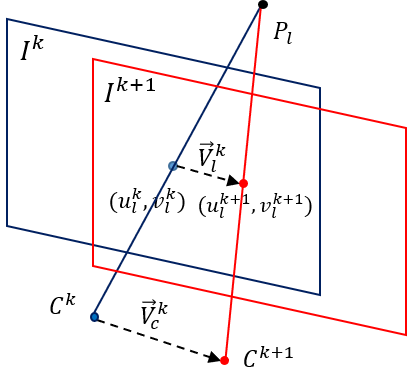}
    \caption{An illustration of feature's velocity on image plane. $I^k$ and $I^{k+1}$ are two consecutive image frames. 
        $[{u}^{k}_l,{v}^{k}_l]$ and $[{u}^{k+1}_l,{v}^{k+1}_l]$ are feature's 2D observations on the image planes $I^k$ and $I^{k+1}$ respectively.
        Camera is assumed to move at a constant speed from $C^k$ to $C^{k+1}$ in short time period $[t_k, t_{k+1}]$.
        Hence, we approximately think that feature $l$ also moves at a constant speed $\mathbf{V}^k_l$ on the image plane in short time period. 
        \label{fig:velocity}}
\end{figure}

As depicted in Fig. \ref{fig:velocity}, $I^k$ and $I^{k+1}$ are two consecutive image frames. 
The camera is assumed to move at a constant speed from $C^k$ to $C^{k+1}$ in the short time period $[t_k, t_{k+1}]$.
Hence, we approximately think that feature $l$ also moves at a constant speed $\mathbf{V}^k_l$ on the image plane in this short time period. 
The velocity $\mathbf{V}^k_l$ is calculated as follows:

\begin{equation}
\label{eq:velocity}
\mathbf{V}^k_l = ({
\begin{bmatrix}
{u}^{k+1}_l \\
{v}^{k+1}_l 
\end{bmatrix} - 
\begin{bmatrix}
{u}^{k}_l \\
{v}^{k}_l 
\end{bmatrix}
})/({t_{k+1} - t_k})
\end{equation}
where $[{u}^{k}_l,{v}^{k}_l]$ and $[{u}^{k+1}_l,{v}^{k+1}_l]$ are feature's 2D observations on the image planes $I^k$ and $I^{k+1}$ respectively.

\subsection{Vision Factor with Time Offset}
\label{sec:vision_factor}
In classical sparse visual SLAM algorithms, visual measurements are formulated as (re)projection error in cost function.
We refector the classical (re)projection error by adding a new variable, time offset.
There are two typical parameterizations of a feature. 
Some algorithms parameterize feature as its 3D position in the global frame, while other algorithms parameterize feature as depth or inverse depth with respect to a certain image frame.
In the following, we respectively model time offset into vision factors with these two kinds of parameterizations. 

\subsubsection{3D Position Parameterization}
The feature is parameterized as 3D position $(\mathbf{P}_l = [x_l, y_l, z_l]^T)$ in the global frame. 
In traditional, the visual measurement is formulated as the projection error,
\begin{equation}
\begin{split}
\mathbf{e}^k_l &= \mathbf{z}^k_l - \pi(\mathbf{R}^{w^T}_{c_k} (\mathbf{P}_l - \mathbf{p}^{w}_{c_k}))\\
\mathbf{z}^k_l &= 
[{u}^{k}_l \ {v}^{k}_l]^{T}.
\end{split}
\end{equation}
$\mathbf{z}^k_l$ is the observation of feature $l$ in frame $k$.
($\mathbf{R}^w_{c_k}, \mathbf{p}^w_{c_k}）$) is the camera pose, which transform feature $\mathbf{P}_l$ from global frame to local camera frame.
$\pi(\cdot)$ denotes the camera projection model, which projects 3D feature into image plane with distortion. 
%$\mathbf{R}^w_{c_k}, \mathbf{p}^w_{c_k}）$ and  $\mathbf{P}_l$ are unknown variables, which need to be optimized. 

The camera pose ($\mathbf{R}^w_{c_k}, \mathbf{p}^w_{c_k}）$) is constrained by visual measurements in the above-mentioned formulation. 
It is also constrained by IMU measurements. 
In practice, if there exists time misalignment between IMU and camera, the IMU constraint is inconsistent with vision constraint in the time domain. 
In other words, we should shift camera sequence forward or backward, so that the camera and IMU data streams become temporally consistent.
Instead of shifting whole camera or IMU sequence, we specifically shift feature' observations in the timeline. 
The new formulation is written as follows,
\begin{equation}
\label{eq:projection_td}
\begin{split}
&\mathbf{e}^k_l =  \mathbf{z}^k_l(t_d)- \pi(\mathbf{R}^{w^T}_{c_k} (\mathbf{P}_l - \mathbf{p}^w_{c_k}))\\
&\mathbf{z}^k_l(t_d) =[{u}^{k}_l \ {v}^{k}_l]^T + t_d \mathbf{V}^k_l.
\end{split}
\end{equation}
$\mathbf{V}^k_l$ is feature's speed on the image plane, got from eq. \ref{eq:velocity}.
$t_d$ is the unknown variable of time offset, which shifts feature's observation in time domain.
By optimizing $t_d$, we can find the optimal camera pose and feature's observation in the time domain which matches IMU constraints.

\begin{figure}
    \centering
    \includegraphics[width=0.4\textwidth]{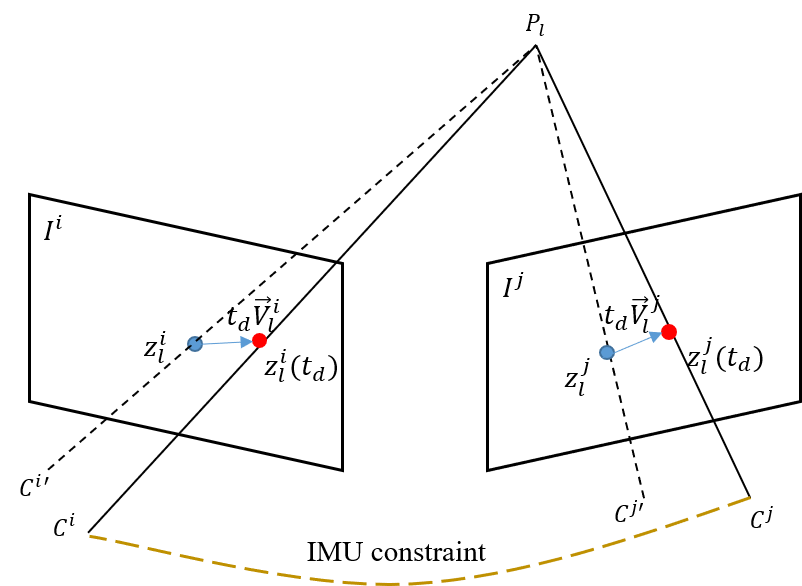}
    \caption{ 
        \label{fig:timeoffset_tri}
        An illustration of reprojection process.
        The dashed line presents traditional reprojection procedure without time offset modeling. 
        The solid line presents proposed reprojection which takes time offset into consideration. 
        The yellow line presents IMU constraint.
        The IMU constraint is inconsistent with traditional reprojection constraint.
    By optimizing $t_d$, we can find the optimal camera pose and feature's observation in time domain which matches IMU constraint.}
\end{figure}

\subsubsection{Depth Parameterization}
The feature can be also parameterized as depth or inverse depth with respect to an image frame.
We take depth $\lambda_i$ in image $i$ as the example.
The traditional reprojection error from image $i$ to image $j$ is written as,
\begin{equation}
\begin{split}
\mathbf{e}^{j}_l &= \mathbf{z}^j_l - \pi(\mathbf{R}^{w^T}_{c_j} ({\mathbf{R}^{w}_{c_i}} \lambda_i 
\begin{bmatrix}
\mathbf{z}^i_l \\ 1
\end{bmatrix}
+ \mathbf{p}^w_{c_i}  - \mathbf{p}^w_{c_j}))\\
\mathbf{z}^i_l &= 
[{u}^{i}_l \ {v}^{i}_l]^{T},  \ \ \ \mathbf{z}^j_l = 
[{u}^{j}_l \ {v}^{j}_l]^{T} .
\end{split}
\end{equation}
The feature $l$ is first projected into global frame, then back projected onto the image plane in local camera frame $j$. 
The residual is the displacement between observation and back projection location.

Similarly with eq. \ref{eq:projection_td}, we take the time offset variable $t_d$ into account,
\begin{equation}
\begin{split}
\mathbf{e}^{j}_l &= \mathbf{z}^j_l(t_d) - \pi(\mathbf{R}^{w^T}_{c_i} ({\mathbf{R}^{w}_{c_j}} \lambda_i 
\begin{bmatrix}
\mathbf{z}^i_l(t_d) \\ 1
\end{bmatrix}
+ \mathbf{p}^w_{c_i}  - \mathbf{p}^w_{c_j}))\\
\mathbf{z}^i_l &= 
[{u}^{i}_l \ {v}^{i}_l]^{T} + t_d \mathbf{V}^i_l,  \ \ \ \mathbf{z}^j_l = 
[{u}^{j}_l \ {v}^{j}_l]^{T} + t_d \mathbf{V}^j_l .
\end{split}
\end{equation}
Fig. \ref{fig:timeoffset_tri} depicts the reprojection process.
The dashed line represents traditional reprojection procedure without time offset modeling. 
The solid line represents proposed reprojection which takes time offset into consideration. 
The yellow line denotes IMU constraint. 
The IMU constraint is inconsistent with traditional reprojection constraint.
By optimizing $t_d$, we can find the optimal camera pose and feature's observation in the time domain which matches IMU constraints.

\subsection{Optimization with Time Offset}
\begin{figure}
    \centering
    \includegraphics[width=0.48\textwidth]{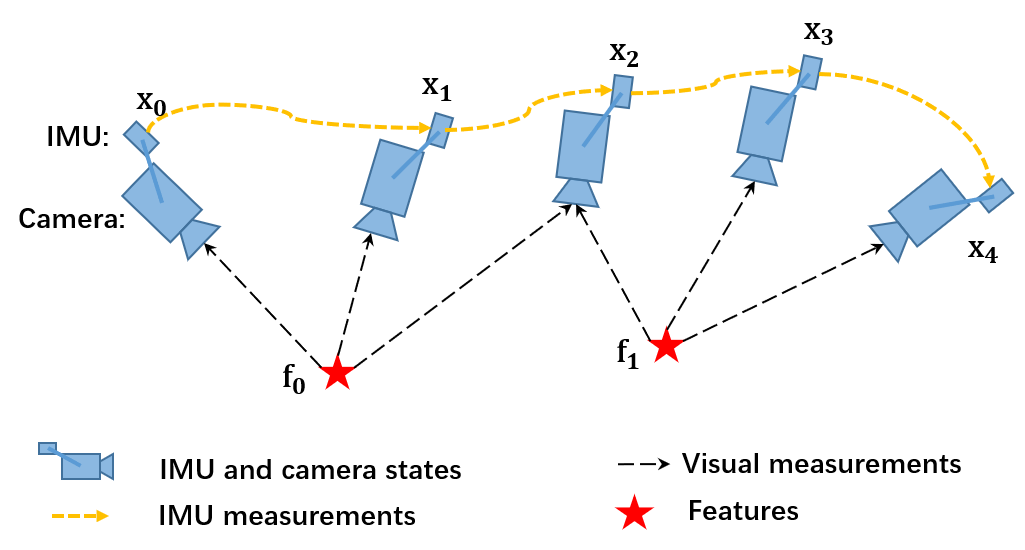}
    \caption{ 
        \label{fig:vins}
        An illustration of visual-inertial localization and mapping problem.
        We maintain several camera frames and IMU measurements in a bundle.
        The bundle size usually is limited to reduce computation complexity.  
        A local bundle adjustment (BA) jointly optimizing camera and IMU states, as well as feature locations. }
\end{figure}

By leveraging the above-mentioned vision factor, we can easily add the temporal calibration function into typical visual-inertial optimization-based frameworks, such as\cite{SheMicKum1505,LeuFurRab1306,qin2017vins}. 
In these frameworks, Visual-inertial localization and mapping is formulated as a nonlinear optimization problem that tightly couples visual and inertial measurements. 
As depicted in Fig. \ref{fig:vins}, several camera frames and IMU measurements are kept in a bundle.
The bundle size usually is limited to bound computational complexity.  
A local bundle adjustment (BA) jointly optimizes camera and IMU states, as well as feature locations.

We can easily add the proposed visual factor (\ref{sec:vision_factor}) into this kind of framework.
To be specific, the whole state variables are augmented with time offset, which are defined as:
\begin{equation}
\label{eq:variable}
\begin{split}
\mathcal{X}    &= \left [ \mathbf{x}_0,\,\mathbf{x}_{1},\, \cdots \,\mathbf{x}_{n},\, \mathbf{P}_0,\,\mathbf{P}_1,\, \cdots \,\mathbf{P}_l,\,t_d \right ] \\
\mathbf{x}_k   &= \left [ \mathbf{p}^w_{k},\,\mathbf{v}^w_{k},\,\mathbf{R}^w_{k}, \,\mathbf{b}_a, \,\mathbf{b}_g \right ], k\in [0,n].
\end{split}
\end{equation}
where the $k$-th IMU state consists of the position $\mathbf{p}^{w}_{k}$, velocity $\mathbf{v}^{w}_{k}$, orientation $\mathbf{R}^{w}_{k}$ in the global frame, and IMU bias $\mathbf{b}_a$, $\mathbf{b}_g$ in the local body frame. 
The feature $\mathbf{P}_l$ is parameterized by either 3D position in the global frame or depth with respect to a certain image frame.

The whole problem is formulated as one cost function containing IMU propagation factor, reprojection factor, as well as a certain prior factor.
Hereby, we use the proposed vision (\ref{sec:vision_factor}) factor to achieve time offset calibration,
\begin{equation}
\begin{aligned}
\label{eq:jointly optimize loop}
\min_{\mathcal{X}}  \left\{ 
\underbrace{\left\| \mathbf{e}_p - \mathbf{H}_p \mathcal{X} \right\|^2}_{\text{prior factor}}
+ \underbrace{\sum_{k \in \mathcal{B}}  \left\| \mathbf{e}_{\mathcal{B}}({\mathbf{z}}^{k}_{{k+1}}, \mathcal{X})  \right\|_{\mathbf{P}^{k}_{{k+1}}}^2}_{\text{IMU propagation factor}}
 \right. \\
\left. 
+ \underbrace{\sum_{(l,j) \in \mathcal{C}}  \left\| \mathbf{e}_{\mathcal{C}}({\mathbf{z}}^{j}_l, \mathcal{X})  \right\|_{\mathbf{P}^{j}_l}^2}_{\text{proposed vision factor}}    \right\}.
\end{aligned}                    
\end{equation}
$\mathbf{e}_{\mathcal{B}}({\mathbf{z}}^{k}_{{k+1}},\, \mathcal{X})$ is the error term from IMU propagation.
$\mathcal{B}$ is the set of all IMU measurements.
$\mathbf{e}_{\mathcal{C}}({\mathbf{z}}^{j}_l,\, \mathcal{X})$ is the proposed visual (re)projection error, which includes the time offset variable.  
$\mathcal{C}$ is the set of features which have been observed at least twice in the image frames. 
The errors are weighted by their inverse covariance $\mathbf{P}$. 
$\{\mathbf{e}_p,\,\mathbf{H}_p\}$ is the prior information from prior knowledge and marginalization.
Only a small amount of measurements and states are kept in the optimization bundle, while others are marginalized out and converted into prior.
The non-linear least squares cost function can be efficiently optimized using Gauss-Newton methods.

\subsection{Compensation of Time Offset}
After each optimization, we compensate time offset by shifting timestamps of subsequent visual streams, as $t_{cam}' = t_{cam} + t_d$. 
Then the system estimates $\delta t_d$ between compensated visual measurement and inertial measurement in the following. 
$\delta t_d$ will be iteratively optimized in subsequent data streams, which will converge to zero. 
With the decrease of time interval $\delta t_d$, our basic assumption (feature moves at a constant speed on the image plane in a short time interval) is more and more reasonable.
Even if there is a huge time offset (e.g. hundreds of milliseconds) at the beginning, the process will compensate it from coarse to fine gradually.

\section{Experiment Results}
\label{sec:experiments}
In this section, we first demonstrate the accuracy of temporal calibration, then we show the overall VIO performance improved by temporal calibration. 
The calibration experiments are presented with simulated data and real sensor set. 
The overall VIO performance is shown with public dataset and real-world experiment. 
In each part, we compare the proposed algorithm against other popular methods.

\subsection{Implement Details}
We adopt the visual-inertial optimization framework proposed in \cite{qin2017vins}. 
We only add the time offset into the state vector and use the proposed vision factor (Sect. \ref{sec:vision_factor}).
Features are detected by Shi-Tomasi Corner Detector\cite{ShiTom9406} and tracked by KLT tracker \cite{LucKan8108}, while IMU measurements are locally integrated.
%The VIO starts with a robust initialization procedure to guarantee the system can launch under any unknown state or motion. 
Poses, velocities, IMU bias of several keyframes, as well as feature position, are optimized in a local bundle adjustment. 
Only keyframes, which contain sufficient feature parallax with their neighbors, are temporarily kept in the local window. 
Previous keyframes are marginalized out of the window in order to bound computation complexity. 
Ceres Solver~\cite{ceres-solver} is used for solving this nonlinear problem.
The whole system runs in real-time with Intel i7-3770 CPU.

\subsection{Temporal Calibration Results}
\subsubsection{Simulation}
We randomly generate 500 feature points in the 60m x 60m x 60m space. 
Features locations are unknown.
Visible features are projected to the virtual camera subjected to zero-mean Gaussian noise with a standard deviation of 0.5 pixels.
Inertial measurements are also subjected to zero-mean Gaussian noise with standard deviation of $0.01m/ s^{2}$ and $0.001 rad/s$ in accelerometer and gyroscope respectively without bias.
The inertial measurement rate is 100 Hz and the camera frame rate is 10 Hz.
The camera and IMU move together with sufficient accelerating and rotating motion.  
The whole trajectory lasts 30 seconds.
We set time offsets as 5ms, 15ms, and 30ms.
For each time offset, 100 simulated trials are conducted.
The calibration results are shown in Table. \ref{tab:simulaiton_calib}.
Our algorithm can successfully calibrate time offset in simulated data with low RMSE (Root Mean Square Error).

\begin{table}
    \centering
    \caption{{Simulation Calibration Results} \label{tab:simulaiton_calib}}
    \begin{tabular}{c|ccc}
        \toprule
        Sequence[ms] & Mean[ms] & RMSE[ms] & NEES  \\
        \midrule
        I. 5          & 5.12  & 0.36 & 7.2\%  \\
        II. 15          & 15.06  & 0.61 & 4.1\%  \\
        III. 30        & 30.17  & 0.68 & 2.3\%  \\
        \bottomrule
    \end{tabular}
 \begin{tablenotes}
    \footnotesize
    \item[1]  The calibration results of simulated data with 5ms, 15ms and 30ms time offset. RMSE is the root mean square error. NEES is normalized estimation error squared.
\end{tablenotes}
\end{table}

\subsubsection{Real Sensor}
\label{subsubsec:rs_calib}

\begin{figure}
    \centering
    \includegraphics[width=0.45\textwidth]{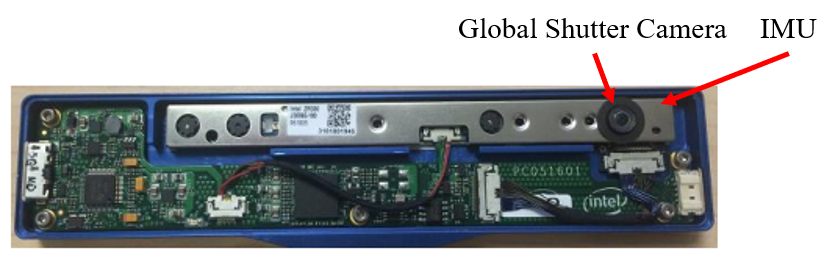}
    \caption{Intel Realsense camera ZR300, which contains a fisheye global shutter camera with $100^\circ \times 133^\circ$ FOV and IMU (Gyro \& Acc).
        \label{fig:realsense} }
\end{figure}

We used Intel Realsense camera ZR300\footnote{https://software.intel.com/en-us/realsense/zr300}, which contains a fisheye global shutter camera with $100^\circ \times 133^\circ$ FOV and IMU (Gyro \& Acc), as shown in Fig. \ref{fig:realsense}. 
The manufacturer claimed that sensors are well synchronized. 
In practice, we find out there is an apparent time offset between the fisheye camera and IMU got from default SDK, which is affected by exposure time.
Since ground truth is unavailable, we take the state-of-art temporal calibration toolbox, Kalibr\cite{furgale2013unified} for comparison. 
Kalibr calibrated time offset in an offline batch optimization framework, which needs addition calibration pattern (chessboard).

We set five exposure times from 20ms to 30ms. 
For each exposure time, we collected fifteen datasets by moving the sensor in front of a calibration chessboard for 40 seconds.
Actually, Kalibr relies on a calibration pattern, while our algorithm does not.
To make the calibration fully observable, we ensured sufficient rotational and accelerated motion over all datasets.

\begin{figure}
    \centering
    \includegraphics[width=0.45\textwidth]{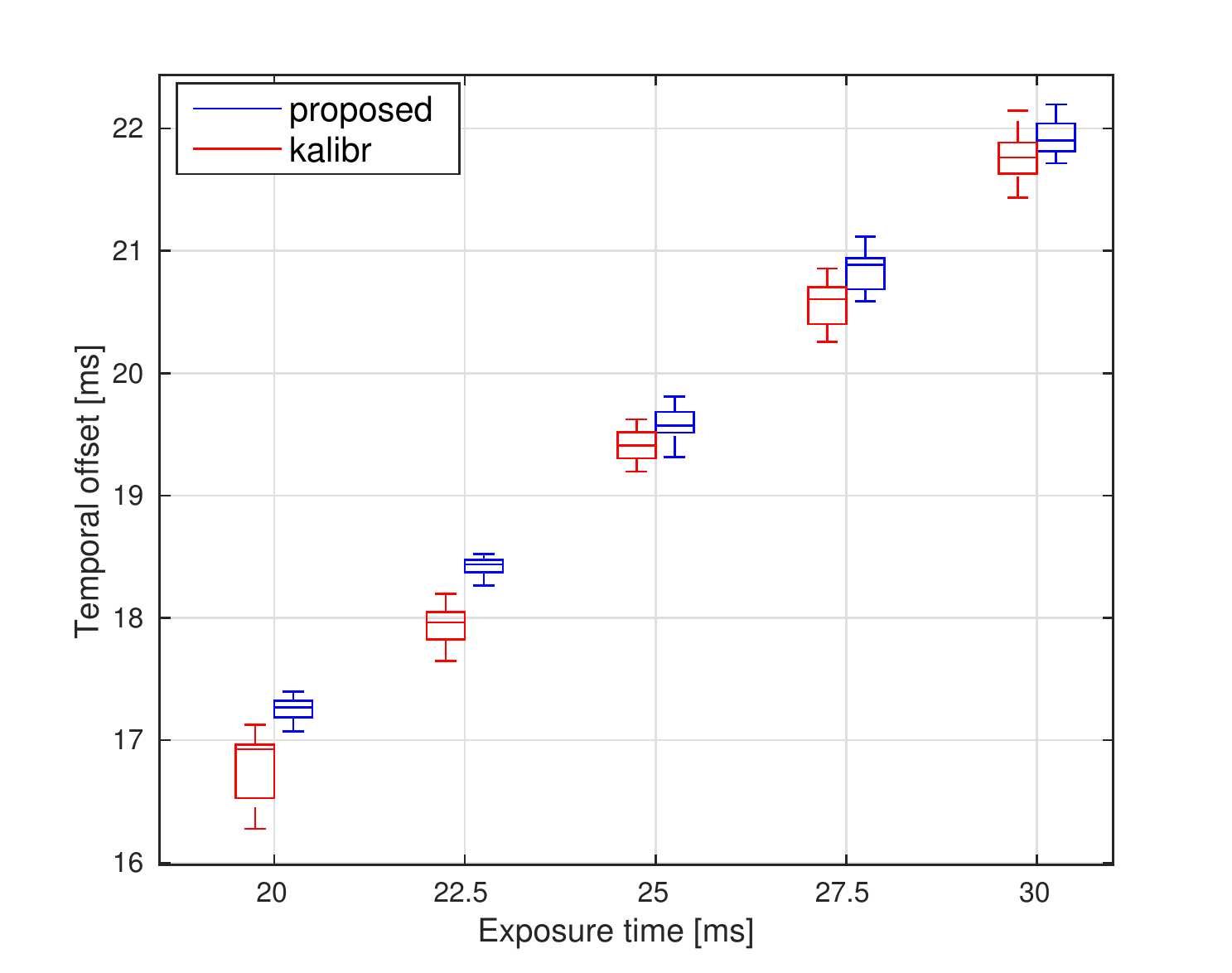}
    \caption{The estimated time offset from the proposed method and Kalibr with respect to different exposure times.
        \label{fig:realsense_calib} }
\end{figure}

The results of temporal calibration are depicted in Fig. \ref{fig:realsense_calib}. 
We can see that the temporal offset evolves linearly with the exposure time with a slope around 0.5.
That is because the middle of the exposure time is treated as the optimal point to timestamp an image.
Therefore, the temporal offset consists of fixed communication and triggering delays plus half exposure time. 
Both the proposed method and Kalibr satisfy this situation.
Since ground truth is unavailable, we take Kalibr's results as reference. 
Our results are quite close to Kalibr's results. 
As for standard derivation, the proposed method achieved [0.095, 0.071, 0.14, 0.16, 0.15], which is smaller than Kalibr's standard derivation [0.27, 0.16, 0.13, 0.18, 0.20]. 
The proposed method outperform Kalibr in terms of consistency. 
Note that Kalibr is an offline batch optimization, which consumes dozens of times more than the proposed method. 
Furthermore, Kalibr relies on calibration pattern.
Hence, the proposed method also outperforms Kalibr in terms of efficiency and practicability.

\subsection{Overall VIO Performance}
\subsubsection{Dataset}

We evaluate the proposed method using EuRoC MAV Visual-Inertial Datasets~\cite{Burri25012016}. 
Datasets are collected onboard a micro aerial vehicle, 
which contains stereo images (Aptina MT9V034 global shutter, WVGA monochrome, 20 FPS), 
synchronized IMU measurements (ADIS16448, 200 Hz), 
and ground truth states (VICON and Leica MS50). 
We only use images from the left camera. 
It is well known that images and IMU measurements are strictly synchronized in this dataset.
To demonstrate the capability of temporal calibration, we set the time offset by manually shifting IMU timestamps. 
To be specific, we add a fixed millisecond value to IMU timestamps, such that there is a fixed time offset between IMU and camera.  
We made time-shifted sequences and used them to test the proposed algorithm and other methods. 

\begin{figure}
	\centering
	\includegraphics[width=0.5\textwidth]{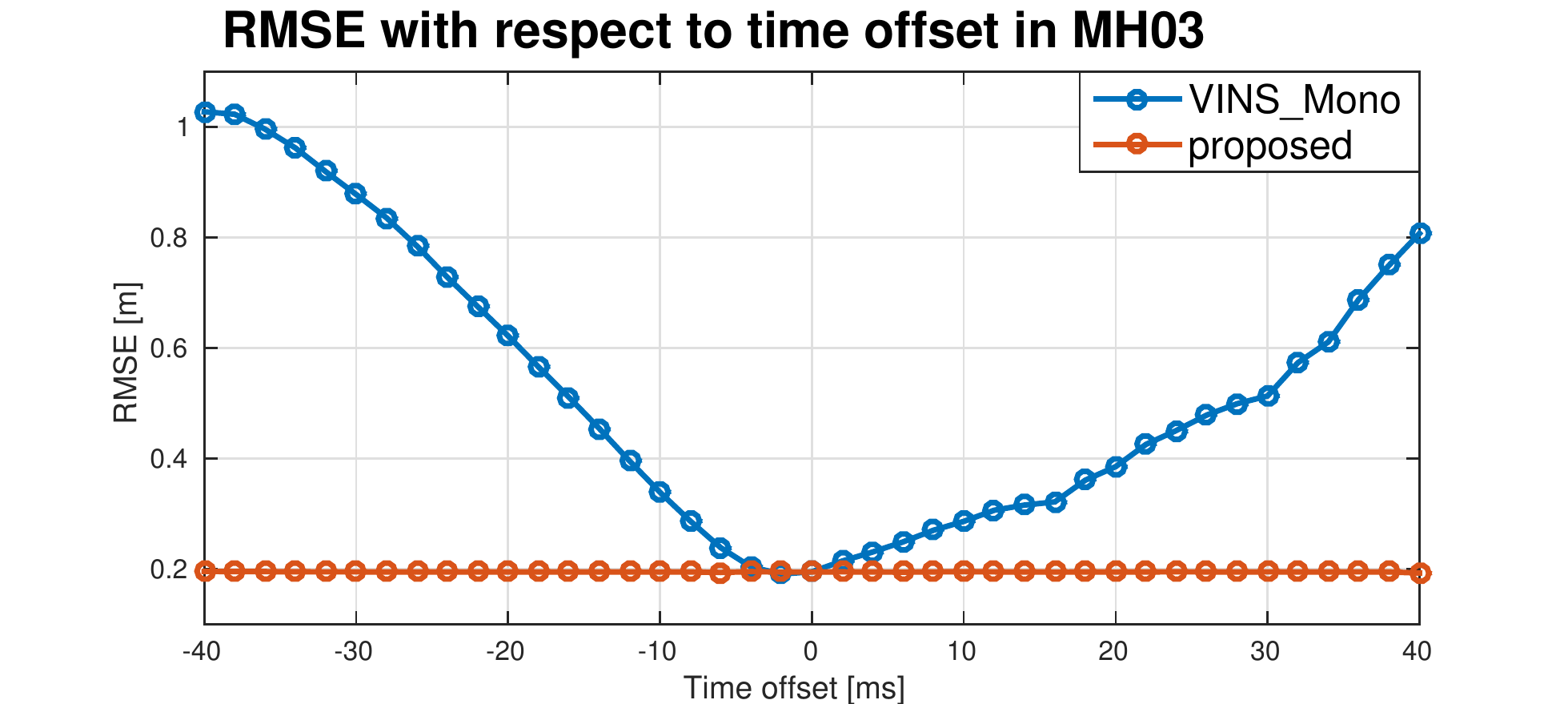}
	\caption{ 
		RMSE with respect to the time offset in MH03 sequence.
		The x-axis shows the predefined time offset, and the y-axis shows the RMSE (Root Mean Square Error)\cite{sturm2012benchmark}.   
		The blue line represents results of VINS-Mono \cite{qin2017vins}, which is the base framework proposed method build on.
		The red line represents results of proposed method, which has capability of time offset calibration.
		\label{fig:RMSE_timeoffset} }
	\vspace{-2cm}
\end{figure}

At first, we studied the influence of time offset on visual-inertial odometry.
We set time offsets from $-40$ to $40$ms, and test these time-biased sequences with VINS-Mono \cite{qin2017vins} and the proposed method respectively.
VINS-Mono is the base framework which we build our system on.
VINS-Mono does not have time offset calibration capability, thus it significantly suffers from temporal misalignment.
The result is depicted in Fig. \ref{fig:RMSE_timeoffset}.
The x-axis shows the predefined time offset, and the y-axis shows RMSE (Root Mean Square Error), as proposed in \cite{sturm2012benchmark}.   
The test data is the MH03 sequence, whose IMU timestamp is shifted.
The blue line represents results of VINS-Mono.
We can see that the RMSE evolves along a parabolic curve with respect to time offsets for VINS-Mono. 
The performance deteriorates dramatically when the time offset increases. 
The tolerance interval is only within 6 milliseconds. 
That demonstrates that it is necessary to do time offset calibration.
The red line represents results of the proposed method, which calibrates the time offset.
It can be seen that the RMSEs are same under different time offsets, which proves that our calibration procedure is very effective.

\begin{table}
	\centering
	\caption{{RMSE in EuRoC dataset.} \label{tab:euroc}}
	\begin{tabular}{c c c c c}
		\toprule
		\multirow{2}{*}{Sequences} & \multirow{2}{*}{$t_d$[ms]} &
		\multicolumn{2}{c}{Proposed} &
		{OKVIS } \\
		\cline{3-4}
		&     & Esti. $t_d$ [ms] & RMSE[m] &  RMSE[m] \\
		\hline
		\multirow{4}{*}{MH\_01}
		& 5 & 4.87 & 0.155 & 0.318 \\
		& 15 & 14.87 & 0.158 & 0.382 \\
		& 30 & 29.87 & 0.156 & 0.544 \\
		\hline
		\multirow{4}{*}{MH\_03}
		& 5 & 4.99 & 0.194 & 0.284 \\
		& 15 & 15.02 & 0.194 & 0.451 \\
		& 30 & 29.99 & 0.195 & 2.805 \\
		\hline
		\multirow{4}{*}{MH\_05}
		& 5 & 5.10 & 0.303 & 0.432 \\
		& 15 & 15.30  & 0.326  & 0.577 \\
		& 30 & 30.08 & 0.303 & 0.652 \\
		\hline
		\multirow{4}{*}{V1\_01}
		& 5 & 5.16 & 0.088 & 0.100 \\
		& 15 & 15.16 & 0.088 & 0.202 \\
		& 30 & 30.21 & 0.089 & 0.257 \\
		\hline
		\multirow{4}{*}{V1\_03}
		& 5 & 4.87 &0.185  &0.349  \\
		& 15 & 14.88  &0.187  & 1.008 \\
		& 30 & 29.90 &0.189  & 1.817  \\
		\hline
		\multirow{4}{*}{V2\_02}
		& 5 & 4.92 &0.159  & 0.378  \\
		& 15 & 14.93 &0.161   & 0.520  \\
		& 30 & 29.92 &0.162  &1.010  \\
		\bottomrule
	\end{tabular}
	\begin{tablenotes}
		\footnotesize
		\item[1]   RMSE is root mean square error, as proposed in \cite{sturm2012benchmark}.
	\end{tablenotes}
\end{table}

\begin{figure}
	\centering
	\includegraphics[width=0.5\textwidth]{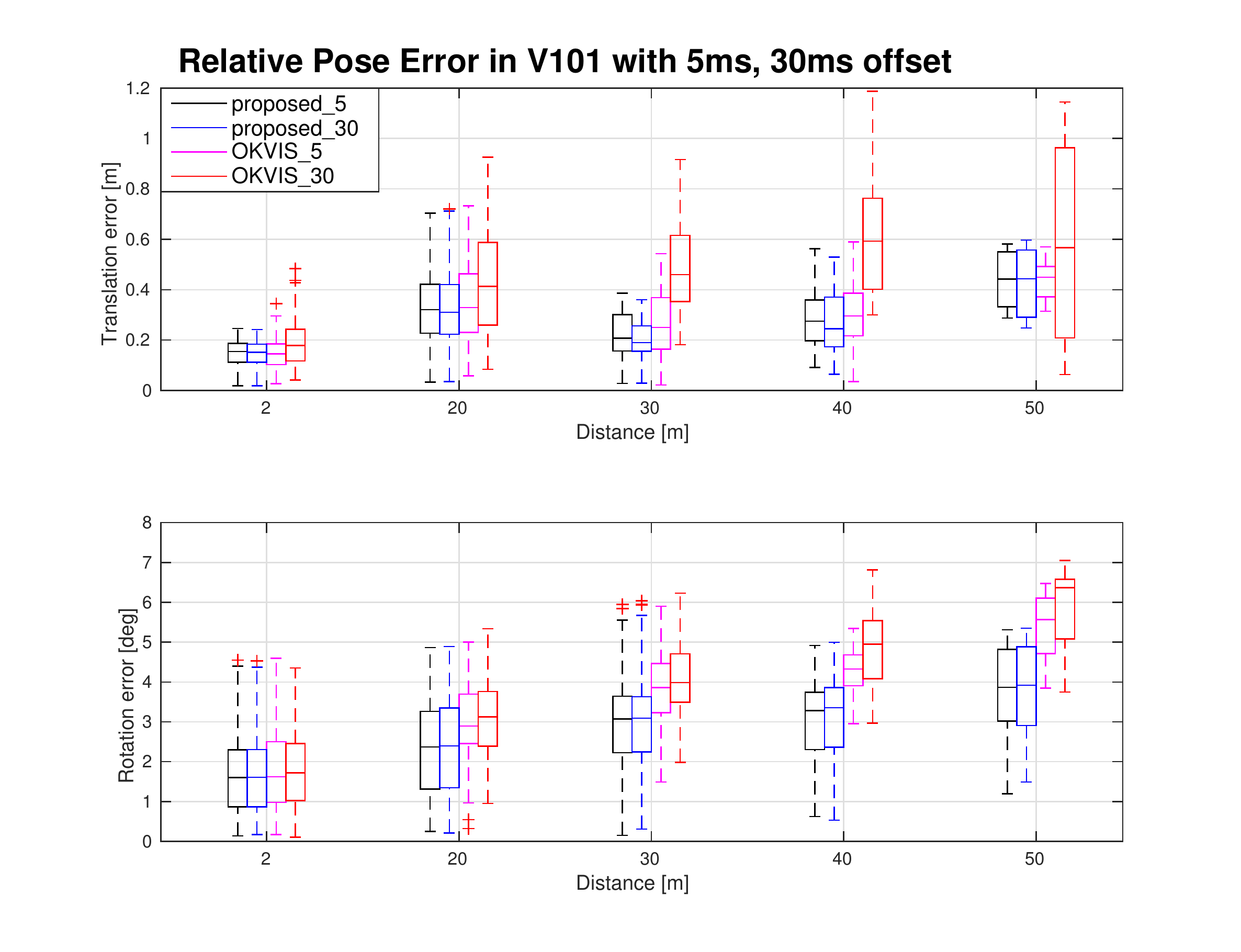}
	\caption{Relative pose error \cite{geiger2012we} comparison between proposed method and OKVIS in V101 sequence with 5ms and 30ms temporal offset. The relative pose errors of proposed method are almost same under two different temporal offsets (black and blue plots). 
		However, the relative pose error of OKVIS increase a lot when temporal offset increases (pink and red plots).
		\label{fig:euroc_detail} }
\end{figure}

\begin{figure}
	\centering
	\includegraphics[width=0.5\textwidth]{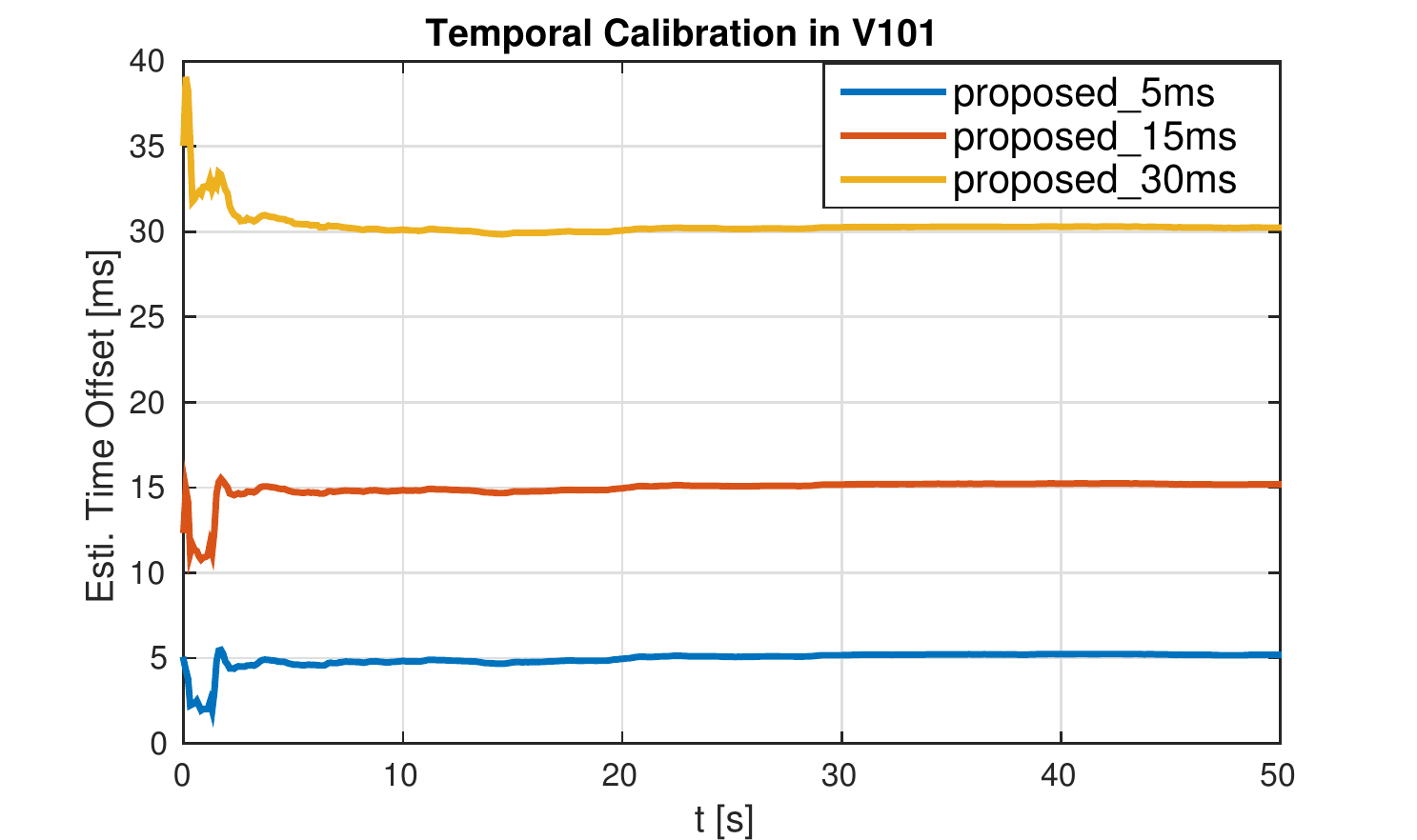}
	\caption{Temporal offset estimation in V101 sequence with 5ms, 15ms and 30ms temporal offset. Estimated offsets converge to the stable value quickly within a few seconds. 
		\label{fig:euroc_td} }
\end{figure}

In the following, we compare against OKVIS~\cite{LeuFurRab1306}, which is another state-of-art visual-inertial odometry algorithm without temporal calibration ability. 
We use time-biased sequences to test proposed method and OKVIS.
The results are shown in TABLE. \ref{tab:euroc}. 
The trajectory is also evaluated by RMSE. 
For OKVIS, with the increase of time offset, the performance degrades (RMSE become larger and larger). 
In some sequences (i.e. MH\_03, V1\_03), RMSE dramatically increases when the time offset is up to 30ms. 
Such time offset makes the system diverge.
For proposed method, however, the performance is not affected by the time offset. 
The RMSEs are almost same in one sequence under different time offset, because the proposed method can calibrate time offset correctly. 
The calibration results are also listed in the Table.
The proposed method can accurately calibrate predefined time offset.
The Proposed method obviously outperforms OKVIS when time offset is larger than 10ms.

Specifically, relative pose error \cite{geiger2012we} comparison between proposed method and OKVIS is shown in Fig. \ref{fig:euroc_detail}. 
The figure is conducted on V101 sequence under 5ms and 30ms temporal offset. 
We can see that the relative pose errors of the proposed method are almost same under two different temporal offsets (black and blue plots). 
However, the relative pose error of OKVIS increases a lot when temporal offset increases (pink and red plots). 

The process of temporal offset estimation is shown in Fig. \ref{fig:euroc_td}.
It can be seen that the estimated offset converges to the stable value quickly only within a few seconds. 
Online temporal calibration significantly benefits overall performance.

\begin{figure}
	\centering
	\includegraphics[width=0.5\textwidth]{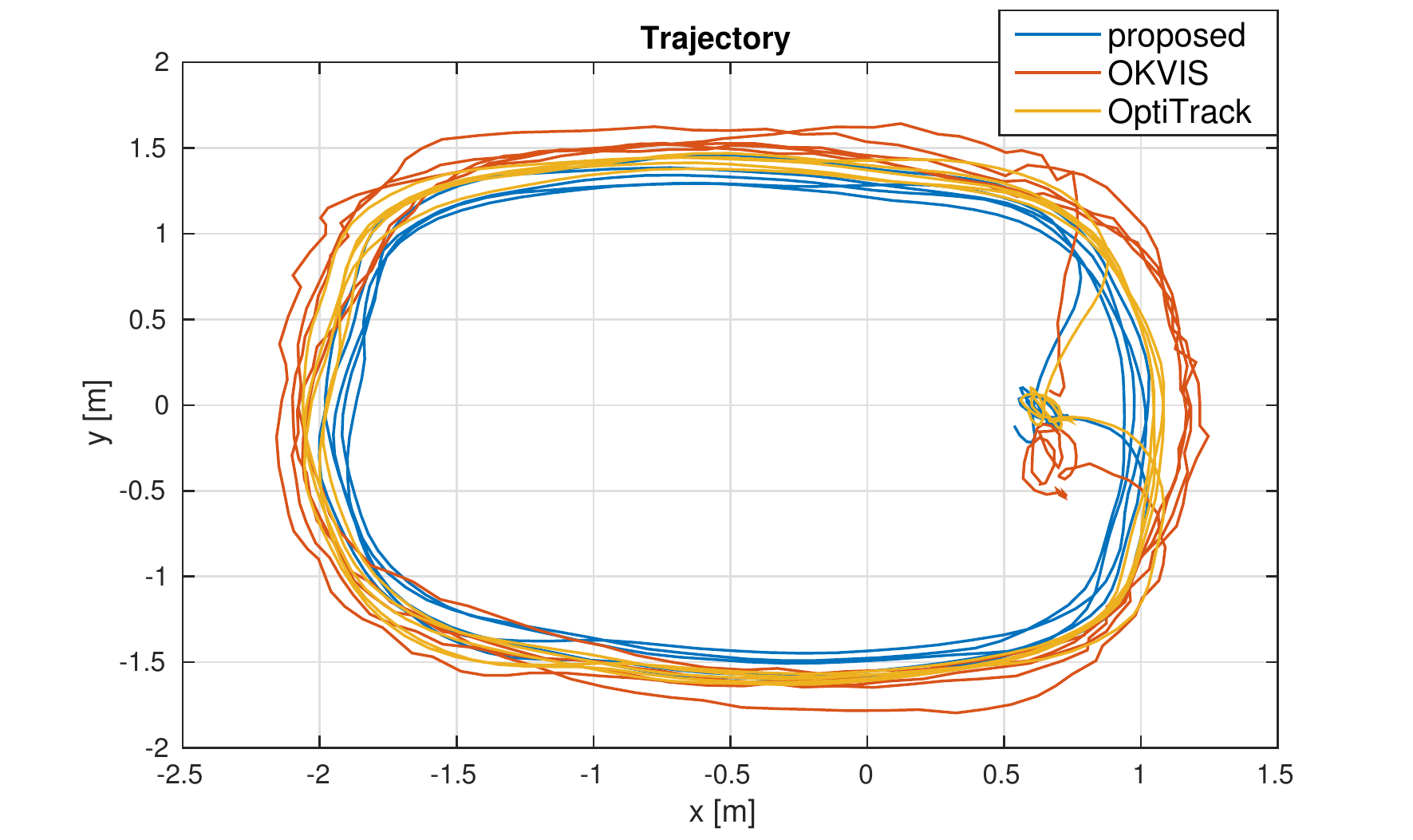}
	\caption{Trajectory of real world experiment. We hold the sensor and walked five circles. Proposed method compares against OKVIS and OptiTrack.
		\label{fig:rs_trajectory} }
\end{figure}

\begin{figure}
	\centering
	\includegraphics[width=0.5\textwidth]{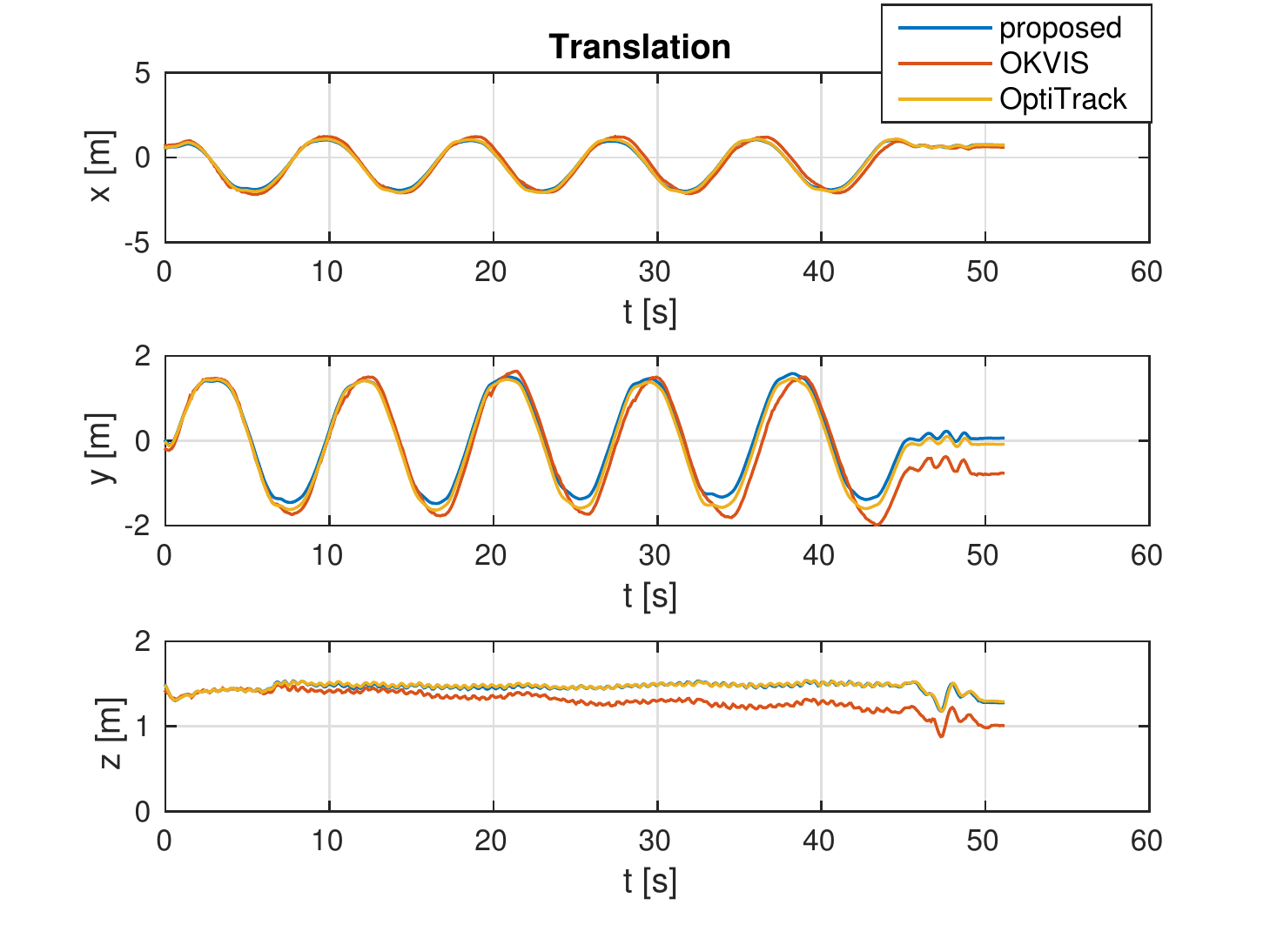}
	\caption{Translation of real world experiment in x, y, and z axis. Proposed method compares against OKVIS and OptiTrack.
		\label{fig:rs_translation} }
\end{figure}

\begin{figure}
	\centering
	\includegraphics[width=0.5\textwidth]{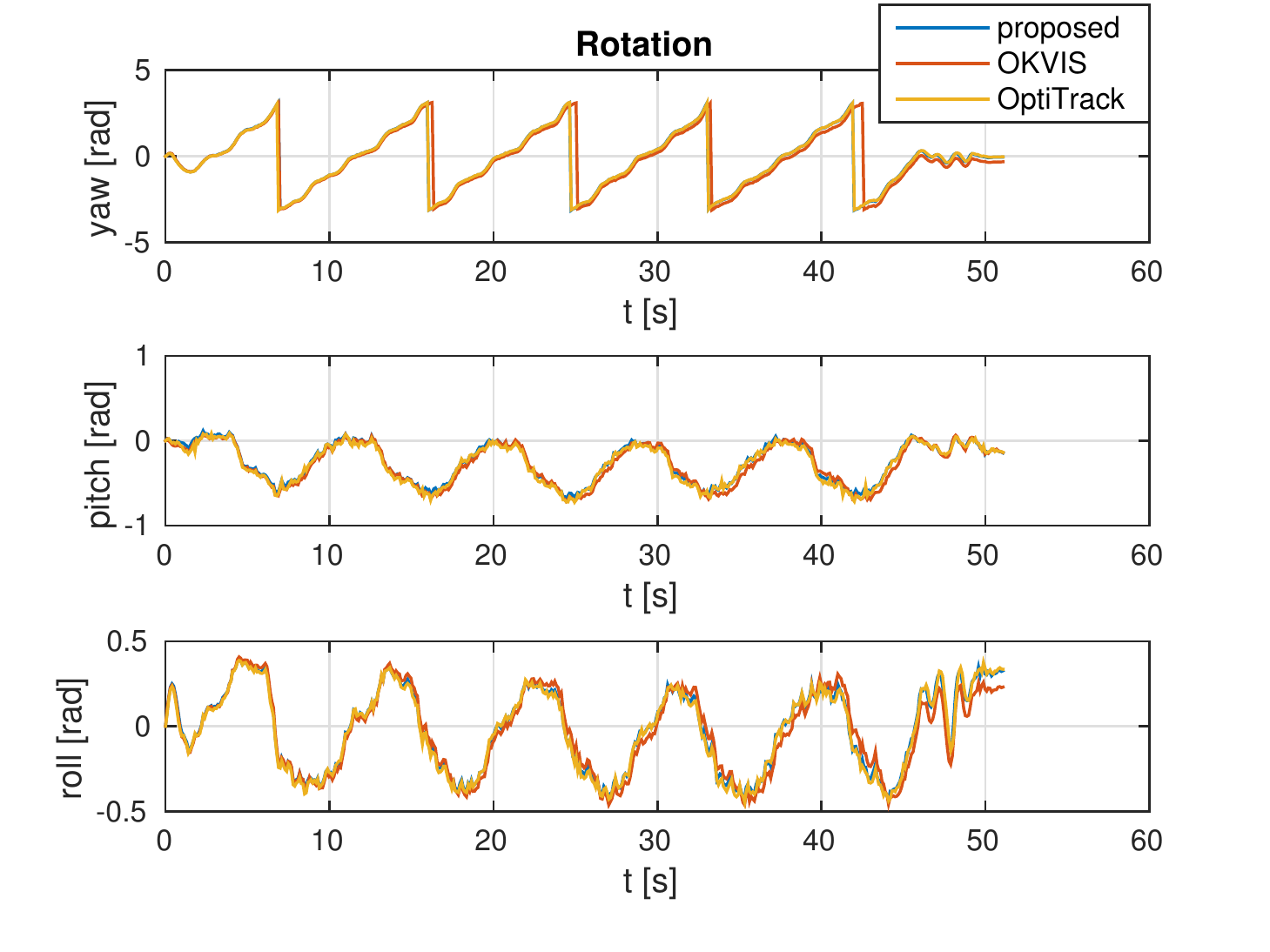}
	\caption{Rotation of real world experiment in yaw, pitch and roll. Proposed method compares against OKVIS and OptiTrack.
		\label{fig:rs_rotation} }
\end{figure}

\begin{figure}
	\centering
	\includegraphics[width=0.5\textwidth]{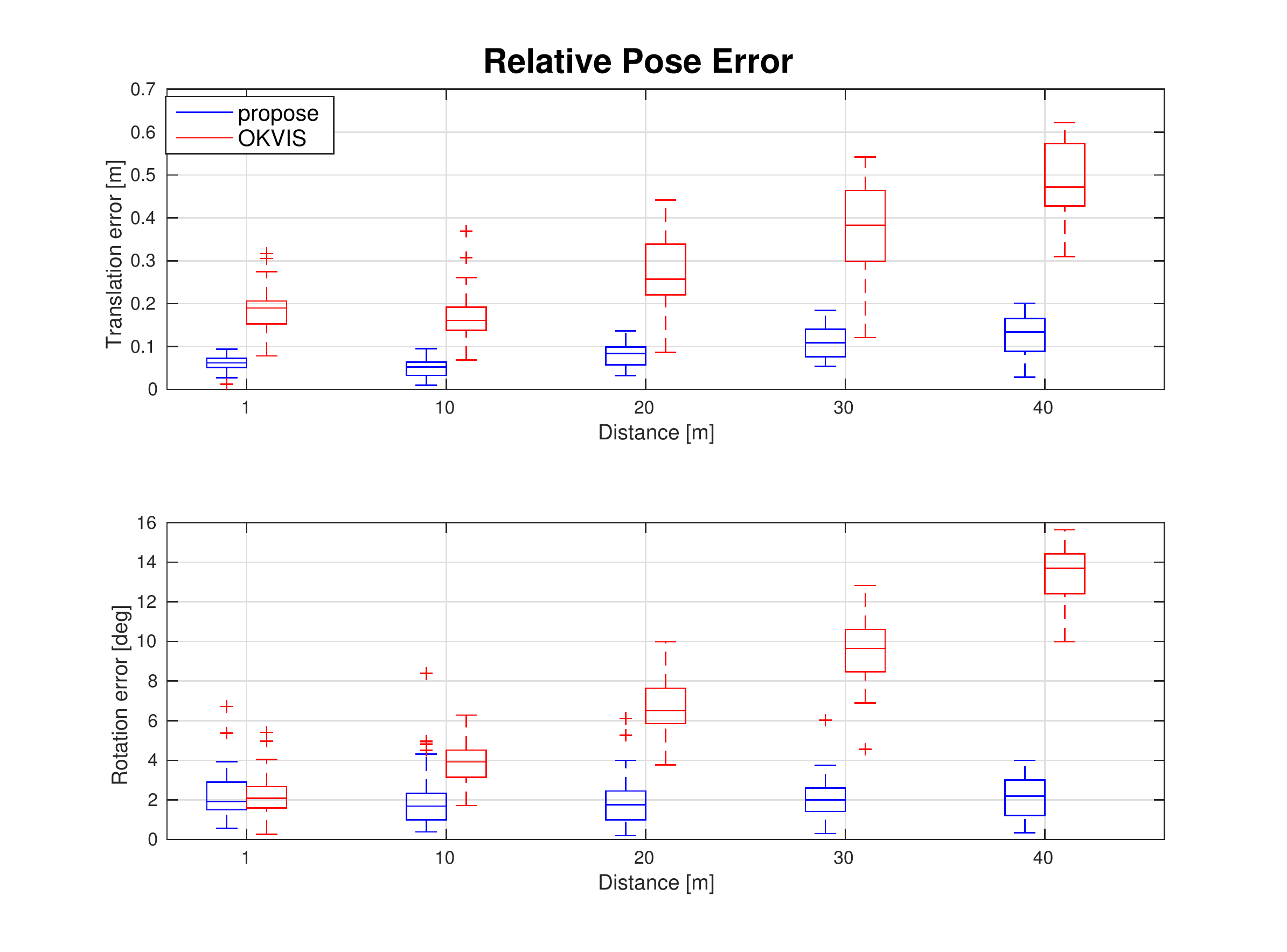}
	\caption{Relative pose error \cite{geiger2012we} comparison between proposed method and OKVIS in real-world experiment. 
		\label{fig:rs_rpe_error} }
\end{figure}

\subsubsection{Real-world Experiment}
We carried out a real-world experiment to validate the proposed system. 
The sensor set is same as in Sec. \ref{subsubsec:rs_calib}, Intel Realsense camera, as shown in Fig. \ref{fig:realsense}. 
The image rate is 30Hz, and the inertial measurement's rate is 350Hz.
We held the sensor suite by hand and walk circularly at a normal pace in a room. 
We compare our results against OKVIS \cite{LeuFurRab1306}. 
Meanwhile, the results from OptiTrack \footnote{http://optitrack.com/} are treated as ground truth.

We held the sensor and walked five circles. 
The trajectory is depicted in Fig. \ref{fig:rs_trajectory}. 
The detailed translation in x, y and z-axis is shown in Fig. \ref{fig:rs_translation}.
The detailed rotation in yaw, pitch, and roll is shown in Fig. \ref{fig:rs_rotation}. 
In translation and rotation comparison, we can see that OKVIS's results drift noticeably along with the time. 
Relative pose error \cite{geiger2012we} comparison between the proposed method and OKVIS is shown in Fig. \ref{fig:rs_rpe_error}. 
The relative pose error of OKVIS is larger than the proposed method. 
Moreover, the relative pose error increases at a faster speed than the proposed method.
Obviously, the proposed method outperforms OKVIS in both translation and rotation due to online temporal calibration.
The temporal offset calibrated from the proposed method is 12.74ms, which will significantly affect VIO performances in a long run without effective calibration and compensation.

\section{Conclusion}
\label{sec:conclusion}
In this paper, we have presented an online approach to calibrate time offset between IMU and camera. 
Our approach is a general model, which can be easily adopted in optimization-based visual-inertial frameworks.
The time offset is jointly optimized along with IMU and camera states, as well as features.
Our simulation and experimental results indicate the proposed approach can achieve high accuracy in both time offset calibration and system's motion estimation, even compared with other state-of-art offline methods.
Although we use the monocular sensor suite to showcase our method in this paper, the proposed method can be easily generalized to multi-camera visual-inertial systems.

\balance
\bibliography{paper.bib}

% Generated by IEEEtran.bst, version: 1.14 (2015/08/26)
\begin{thebibliography}{10}
\providecommand{\url}[1]{#1}
\csname url@samestyle\endcsname
\providecommand{\newblock}{\relax}
\providecommand{\bibinfo}[2]{#2}
\providecommand{\BIBentrySTDinterwordspacing}{\spaceskip=0pt\relax}
\providecommand{\BIBentryALTinterwordstretchfactor}{4}
\providecommand{\BIBentryALTinterwordspacing}{\spaceskip=\fontdimen2\font plus
\BIBentryALTinterwordstretchfactor\fontdimen3\font minus
  \fontdimen4\font\relax}
\providecommand{\BIBforeignlanguage}[2]{{%
\expandafter\ifx\csname l@#1\endcsname\relax
\typeout{** WARNING: IEEEtran.bst: No hyphenation pattern has been}%
\typeout{** loaded for the language `#1'. Using the pattern for}%
\typeout{** the default language instead.}%
\else
\language=\csname l@#1\endcsname
\fi
#2}}
\providecommand{\BIBdecl}{\relax}
\BIBdecl

\bibitem{davison2007monoslam}
A.~J. Davison, I.~D. Reid, N.~D. Molton, and O.~Stasse, ``Monoslam: Real-time
  single camera slam,'' \emph{IEEE transactions on pattern analysis and machine
  intelligence}, vol.~29, no.~6, pp. 1052--1067, 2007.

\bibitem{klein2007parallel}
G.~Klein and D.~Murray, ``Parallel tracking and mapping for small ar
  workspaces,'' in \emph{Mixed and Augmented Reality, 2007. IEEE and ACM
  International Symposium on}, 2007, pp. 225--234.

\bibitem{ForPizSca1405}
C.~Forster, M.~Pizzoli, and D.~Scaramuzza, ``{SVO}: Fast semi-direct monocular
  visual odometry,'' in \emph{Proc. of the {IEEE} Int. Conf. on Robot. and
  Autom.}, Hong Kong, China, May 2014.

\bibitem{engel2014lsd}
J.~Engel, T.~Sch{\"o}ps, and D.~Cremers, ``Lsd-slam: Large-scale direct
  monocular slam,'' in \emph{European Conference on Computer Vision}.\hskip 1em
  plus 0.5em minus 0.4em\relax Springer International Publishing, 2014, pp.
  834--849.

\bibitem{mur2015orb}
R.~Mur-Artal, J.~Montiel, and J.~D. Tardos, ``Orb-slam: a versatile and
  accurate monocular slam system,'' \emph{{IEEE} Trans. Robot.}, vol.~31,
  no.~5, pp. 1147--1163, 2015.

\bibitem{kaess2012isam2}
M.~Kaess, H.~Johannsson, R.~Roberts, V.~Ila, J.~J. Leonard, and F.~Dellaert,
  ``isam2: Incremental smoothing and mapping using the bayes tree,'' \emph{Int.
  J. Robot. Research}, vol.~31, no.~2, pp. 216--235, 2012.

\bibitem{engel2017direct}
J.~Engel, V.~Koltun, and D.~Cremers, ``Direct sparse odometry,'' \emph{IEEE
  Transactions on Pattern Analysis and Machine Intelligence}, 2017.

\bibitem{corke2007introduction}
P.~Corke, J.~Lobo, and J.~Dias, ``An introduction to inertial and visual
  sensing,'' \emph{Int. J. Robot. Research}, vol.~26, no.~6, pp. 519--535,
  2007.

\bibitem{MouRou0704}
A.~I. Mourikis and S.~I. Roumeliotis, ``A multi-state constraint {K}alman
  filter for vision-aided inertial navigation,'' in \emph{Proc. of the {IEEE}
  Int. Conf. on Robot. and Autom.}, Roma, Italy, Apr. 2007, pp. 3565--3572.

\bibitem{LiMou1305}
M.~Li and A.~Mourikis, ``High-precision, consistent {EKF}-based visual-inertial
  odometry,'' \emph{Int. J. Robot. Research}, vol.~32, no.~6, pp. 690--711, May
  2013.

\bibitem{weiss2012real}
S.~Weiss, M.~W. Achtelik, S.~Lynen, M.~Chli, and R.~Siegwart, ``Real-time
  onboard visual-inertial state estimation and self-calibration of mavs in
  unknown environments,'' in \emph{Proc. of the {IEEE} Int. Conf. on Robot. and
  Autom.}, 2012, pp. 957--964.

\bibitem{lynen2013robust}
S.~Lynen, M.~W. Achtelik, S.~Weiss, M.~Chli, and R.~Siegwart, ``A robust and
  modular multi-sensor fusion approach applied to mav navigation,'' in
  \emph{Proc. of the {IEEE/RSJ} Int. Conf. on Intell. Robots and Syst.}\hskip
  1em plus 0.5em minus 0.4em\relax IEEE, 2013, pp. 3923--3929.

\bibitem{SheMicKum1505}
S.~Shen, N.~Michael, and V.~Kumar, ``Tightly-coupled monocular visual-inertial
  fusion for autonomous flight of rotorcraft {MAV}s,'' in \emph{Proc. of the
  {IEEE} Int. Conf. on Robot. and Autom.}, Seattle, WA, May 2015.

\bibitem{mur2017visual}
R.~Mur-Artal and J.~D. Tard{\'o}s, ``Visual-inertial monocular slam with map
  reuse,'' \emph{IEEE Robotics and Automation Letters}, vol.~2, no.~2, pp.
  796--803, 2017.

\bibitem{forster2017manifold}
C.~Forster, L.~Carlone, F.~Dellaert, and D.~Scaramuzza, ``On-manifold
  preintegration for real-time visual--inertial odometry,'' \emph{IEEE
  Transactions on Robotics}, vol.~33, no.~1, pp. 1--21, 2017.

\bibitem{LeuFurRab1306}
S.~Leutenegger, S.~Lynen, M.~Bosse, R.~Siegwart, and P.~Furgale,
  ``Keyframe-based visual-inertial odometry using nonlinear optimization,''
  \emph{Int. J. Robot. Research}, vol.~34, no.~3, pp. 314--334, Mar. 2014.

\bibitem{bloesch2015robust}
M.~Bloesch, S.~Omari, M.~Hutter, and R.~Siegwart, ``Robust visual inertial
  odometry using a direct ekf-based approach,'' in \emph{Proc. of the
  {IEEE/RSJ} Int. Conf. on Intell. Robots and Syst.}\hskip 1em plus 0.5em minus
  0.4em\relax IEEE, 2015, pp. 298--304.

\bibitem{ForCarDel1507}
C.~Forster, L.~Carlone, F.~Dellaert, and D.~Scaramuzza, ``{IMU} preintegration
  on manifold for efficient visual-inertial maximum-a-posteriori estimation,''
  in \emph{Proc. of Robot.: Sci. and Syst.}, Rome, Italy, Jul. 2015.

\bibitem{Mair2011Spatio}
E.~Mair, M.~Fleps, M.~Suppa, and D.~Burschka, ``Spatio-temporal initialization
  for imu to camera registration,'' in \emph{IEEE International Conference on
  Robotics and Biomimetics}, 2011, pp. 557--564.

\bibitem{Kelly2010A}
J.~Kelly and G.~S. Sukhatme, ``A general framework for temporal calibration of
  multiple proprioceptive and exteroceptive sensors,'' \emph{Springer Tracts in
  Advanced Robotics}, vol.~79, pp. 195--209, 2010.

\bibitem{furgale2013unified}
P.~Furgale, J.~Rehder, and R.~Siegwart, ``Unified temporal and spatial
  calibration for multi-sensor systems,'' in \emph{Proc. of the {IEEE/RSJ} Int.
  Conf. on Intell. Robots and Syst.}\hskip 1em plus 0.5em minus 0.4em\relax
  IEEE, 2013, pp. 1280--1286.

\bibitem{li20133}
M.~Li and A.~I. Mourikis, ``3-d motion estimation and online temporal
  calibration for camera-imu systems,'' in \emph{Proc. of the {IEEE} Int. Conf.
  on Robot. and Autom.}\hskip 1em plus 0.5em minus 0.4em\relax IEEE, 2013.

\bibitem{qin2017vins}
T.~Qin, P.~Li, and S.~Shen, ``Vins-mono: A robust and versatile monocular
  visual-inertial state estimator,'' \emph{arXiv preprint arXiv:1708.03852},
  2017.

\bibitem{ShiTom9406}
J.~Shi and C.~Tomasi, ``Good features to track,'' in \emph{Proc. of the {IEEE}
  Int. Conf. on Pattern Recognition}, Seattle, WA, Jun. 1994, pp. 593--600.

\bibitem{LucKan8108}
B.~D. Lucas and T.~Kanade, ``An iterative image registration technique with an
  application to stereo vision,'' in \emph{Proc. of the Intl. Joint Conf. on
  Artificial Intelligence}, Vancouver, Canada, Aug. 1981, pp. 24--28.

\bibitem{ceres-solver}
S.~Agarwal, K.~Mierle, and Others, ``Ceres solver,''
  \url{http://ceres-solver.org}.

\bibitem{Burri25012016}
M.~Burri, J.~Nikolic, P.~Gohl, T.~Schneider, J.~Rehder, S.~Omari, M.~W.
  Achtelik, and R.~Siegwart, ``The euroc micro aerial vehicle datasets,''
  \emph{Int. J. Robot. Research}, 2016.

\bibitem{sturm2012benchmark}
J.~Sturm, N.~Engelhard, F.~Endres, W.~Burgard, and D.~Cremers, ``A benchmark
  for the evaluation of rgb-d slam systems,'' in \emph{Proc. of the {IEEE/RSJ}
  Int. Conf. on Intell. Robots and Syst.}, 2012, pp. 573--580.

\bibitem{geiger2012we}
A.~Geiger, P.~Lenz, and R.~Urtasun, ``Are we ready for autonomous driving? the
  kitti vision benchmark suite,'' in \emph{Proc. of the {IEEE} Int. Conf. on
  Pattern Recognition}, 2012, pp. 3354--3361.

\end{thebibliography}

\end{document}